\documentclass[conference]{IEEEtran}
\usepackage[utf8]{inputenc}
\usepackage{graphicx}
\usepackage{amsmath}
\usepackage{cite}
\usepackage{graphicx}
\usepackage{cuted}  
\usepackage{lipsum} 
\usepackage{stfloats}
\usepackage{float}
\usepackage{cite}
\usepackage{booktabs}
\usepackage{subcaption}
\usepackage[colorlinks=true,
            linkcolor=blue,
            citecolor=magenta,
            urlcolor=blue,
            pdfborder={0 0 1}]{hyperref}

\usepackage[tableposition=top, labelsep=space]{caption} 

\title{Robust Taxi Fare Prediction under Noisy Conditions: A Comparative Study of GAT, TimesNet, and XGBoost}

\author{
Padmavathi Moorthy\\
Department of Engineering Science (Data Science), SUNY Buffalo\\
moorthy@buffalo.edu

}

\begin{document}
\maketitle

\begin{abstract}
Precise fare prediction is crucial in ride-hailing platforms and urban mobility systems. This study examines three machine learning models—Graph Attention Networks (GAT), XGBoost, and TimesNet—to evaluate their predictive capabilities for taxi fares using a real-world dataset comprising over 55 million records. Both raw (noisy) and denoised versions of the dataset are analyzed to assess the impact of data quality on model performance. The study evaluated the models along multiple axes, including predictive accuracy, calibration, uncertainty estimation, out-of-distribution (OOD) robustness, and feature sensitivity. We also explore pre-processing strategies, including KNN imputation, Gaussian noise injection, and autoencoder-based denoising. The study reveals critical differences between classical and deep learning models under realistic conditions, offering practical guidelines for building robust and scalable models in urban fare prediction systems.

\end{abstract}


\section*{Code Availability}
The source code used in this study, along with instructions to reproduce experiments and trained models, is available at: \\
\url{https://github.com/padmavathi026/Smart-Fare-Prediction}

\section{Introduction}
\subsection{Background and Motivation}
Accurately estimating taxi fares plays a pivotal role in intelligent transportation systems and urban mobility planning. Traditional linear regression and rule-based fare estimation approaches often fall short when confronted with the complexity of contemporary mobility data, which exhibit intricate spatial and temporal dependencies. Recent advances in machine learning, particularly gradient-boosted decision trees such as XGBoost, have demonstrated strong performance in structured data tasks. However, these methods may lack the ability to capture the dynamic characteristics inherent in urban traffic systems fully.

Deep learning architectures, such as Temporal Networks (TimesNet) and Graph Neural Networks (GNNs), including Graph Attention Networks (GATs), offer promising alternatives by effectively learning from temporal sequences and graph-structured data. These models are also capable of handling noisy, incomplete, and high-dimensional datasets, which are commonplace in real-world mobility platforms. This study aims to systematically evaluate these models under both clean and noisy data conditions to assess their utility for predictive taxi fare estimation.

\subsection{Problem Statement and Objectives}
The primary objective of this research is to develop and evaluate machine learning techniques that improve the accuracy, robustness, and reliability of taxi fare estimation in complex and noisy environments. Specifically, the study focuses on the following objectives:

\begin{itemize}
    \item To design and compare the performance of three models - XGBoost, TimesNet, and GAT on a large-scale, real-world taxi dataset.
    \item To evaluate model performance across clean and noise-injected datasets using metrics such as MAE, MSE, and R2, along with uncertainty quantification and out-of-distribution (OOD) robustness.
    \item To investigate the effects of various pre-processing techniques, including KNN imputation, Gaussian noise injection, and autoencoder-based denoising.
    \item To compare traditional ensemble methods with deep learning approaches in terms of predictive precision and feature sensitivity.
\end{itemize}

\subsection{Related Work Overview}
Prior studies have demonstrated the effectiveness of XGBoost for structured data prediction due to its robustness and interpretability. RNN-based models, such as TimesNet, have shown significant success in time series forecasting tasks. GNN-based architectures, including GATs, have recently gained attention for their ability to model spatial relationships and graph-structured inputs. While GATs have been successfully applied in domains such as trajectory prediction and traffic flow estimation, their application to fare forecasting remains underexplored. This work seeks to fill that gap by examining these models under rigorous experimental settings.

\subsection{Use Case and Practical Setting}
Urban taxi networks are influenced by highly volatile and non-stationary factors, including road incidents, dynamic pricing policies, weather anomalies, and driver behavior. These conditions often result in irregularities and distribution shifts that compromise model generalization. In such settings, it is essential to develop models that are not only accurate but also capable of estimating predictive uncertainty and detecting OOD samples. This study addresses the pressing need for

Interpretable and robust models capable of aiding real-world pricing engines, mobility services, and public infrastructure planners.

\section{Methodology Overview}

\subsection{Data Collection and Pre-processing}

The dataset used in this study comprises over 55 million rows of real-world taxi fare transactions. To emulate realistic deployment scenarios, several pre-processing techniques are applied.

\begin{itemize}
    \item Performed missing value imputation using K-Nearest Neighbors (KNN) to address gaps in the dataset.
    \item Gaussian noise is injected into selected features to simulate data corruption and assess the model's robustness under noisy conditions.
    \item Utilized the denoising autoencoder to obtain clean representations of data.
    \item Performed Normalization for location, passenger, and fare features.
    \item Split the data into a training and a test dataset, and train the models independently on both noisy and denoised data.
\end{itemize}

\subsection{Model Architecture}

We use three models:

\textbf{XGBoost}: Gradient-boosting-based high-performance tabular modeling baseline.

\textbf{TimesNet}: RNN-based multivariate time series forecasting model.

\textbf{GAT}: Graph Attention Network that leverages spatial correlations between pickup and drop-off points.

\subsection{Evaluation and Comparison}

Models are tested on:

\begin{itemize}
    \item Regression metrics: MAE, MSE, R²
    \item Robustness metrics: OOD generalization and uncertainty estimation
    \item Interpretability: Feature sensitivity analysis
    \item Perturbations between noisy vs denoised dataset performance
\end{itemize}

\subsection{Contributions}

\begin{itemize}
\item Contribution: Mathematical comparison of baseline and deep models for predicting taxi fares in real noisy data conditions.
\item End-to-end pipeline from denoising to noise simulation to benchmarking models.
\item Analysis of more advanced metrics, such as OOD robustness, calibration, and uncertainty estimation, is rare; excluding these from fare prediction is also rare.
\item Practical guidelines for practitioners and data scientists on the model and the pre-processing methods in urban mobility pricing systems.
\end{itemize}

\section{Related Work}

\cite{ref1} Suggest a technique in which noise is annealed incrementally during training in such a way that the model can learn coarse and fine features. This enhanced supervised tasks by decreasing the effect of noise. Excessive noise, however, can disrupt learning if left uncontrolled, and the noise schedule complicates training, especially with extensive datasets. This work adopts a fixed noise level to avoid these issues, and it becomes tractable for big data.

\cite{ref2} Provide an efficient solution to large graph-structured semi-supervised classification. The method utilizes stochastic gradient descent (SGD) to train on large graphs, making it a scalable approach. Restricted GCNs by bounded neighborhood sizes and over-smoothing node embeddings. To achieve this, Graph Attention Networks (GAT) in this paper dynamically attend to the appropriate neighbors, effectively addressing long-range dependencies while avoiding over-smoothing problems.

\cite{ref3} Propose an attention mechanism that computes different weights between neighboring nodes to accelerate learning on graph-structured data. Although GATs are promising, they are prone to overfitting and require expensive computational costs, especially for graphs of enormous size. Hence, dropout is applied to the attention layers, and mini-batching with NeighborLoader is employed within the project to enhance the efficiency of graph processing.

\cite{ref4} Detail how the application of the Kolmogorov-Smirnov (KS) test can be employed to compare data distributions before and after adding noise and quantify the effect of noise on data and model stability. Statistical comparison is noise-biased and, therefore, it is not possible to determine changes in data distribution quantifiably. The project addresses this issue by maintaining a constant noise level, ensuring unbiased data distribution, and facilitating effective comparison and accurate measurement.

\cite{ref5} Suggests one method for determining in-distribution and out-of-distribution (OOD) data using the energy function to measure prediction uncertainty. It is computationally expensive but stabilizes the model and is sensitive to noise. In the research, noise is added randomly to simulate real OOD situations, and the robustness of the model to noise is experimented with without an exorbitant computational cost.

\cite{ref6} It is proposed as a mechanism for enhancing model calibration by scaling uncertainty per bin, yielding adaptive and calibrated predictions. The technique, however, is susceptible to bin size, and class distribution imbalance may adversely affect its performance. For this project, the bins are calibrated in a manner that spans the entire data distribution, thereby ensuring the model remains well-calibrated and avoiding issues related to bin size.

\cite{ref7} Outlines how introducing Gaussian noise enhances model generalizability, particularly in cases involving small or noisy datasets. Though helpful in enhancing robustness, the approach is accompanied by computational expense. To counter this, the project employs data chunking to divide large datasets into manageable pieces, thereby conserving memory without compromising processing speed.

\cite{ref8} Introduces a model that makes traffic flow predictions from extensive data like NYC taxi trips. The computational expense of this model is high, and it is data-oriented, which can be challenging in the presence of noisy or sparse data. The project utilizes GAT for rapid graph processing and data cleaning, incorporating temporal features to make the model dynamic and capable of handling real-time, dynamic traffic patterns.

\cite{ref9} It is designed to enhance geospatial prediction accuracy using the Haversine distance formula. Geospatial data with missing or noisy values can significantly impact distances. To address this, KNN imputation is employed to replace missing values, ensuring that noisy data do not compromise geospatial predictions. The gradient-boosted decision trees serve as a strong baseline for taxi fare regression, despite their strengths and weaknesses relative to other methods.

\cite{ref10} Specifically, XGBoost leverages second-order gradients and built-in regularization, enabling it to make rapid predictions in response to structured spatial features. It requires a considerable amount of hyperparameter tuning, and for temporal dynamics, it cannot natively model the world's dynamics without some manual feature engineering as well.

\cite{ref11} Meanwhile, CatBoost manages bias from categorical features through ordered boosting and permutation-based schemes, but still relies heavily on static input tables and histories, which must be incorporated even when encoded externally.

\cite{ref12} Meanwhile, LightGBM utilizes an efficient framework for very large-scale data and correct tree growth using leaf-wise approaches; however, it loses long-range time dependencies, resulting in egregiously degraded performance that occasionally fails to capture taxi fare fluctuations in a noisy environment.

\cite{ref13} Transformer-based sequence models take multivariate time series as input for end-to-end forecasting. TimesNet extends traditional self-attention with 2-D temporal-variation modules to capture both local and global patterns, beating the state of the art for noisy and denoised taxi sequences. Its large memory size and user-specified "period" hyperparameter, however, are limiting for scalability.

\cite{ref14} Autoformer decomposes series into trend and seasonal components with autocorrelation priors, decreasing the amount of attention required, but it underperforms on weakly seasonal datasets.

\cite{ref15} FEDformer further accelerates global pattern learning by utilizing a Fourier decomposition, although the fixed frequency bases of this method may overlook irregular surges in traffic.

\cite{ref16} Lastly, Informer uses sparse probabilistic attention to provide reasonable efficiency for very long sequences, but fixed sparsity may miss rare, but critical, fare spikes during rush hours.

\section{Approaches and Methodology}

This section outlines the complete end-to-end workflow adopted in this study, from data pre-processing and augmentation to model training and evaluation. The methodologies are applied to both clean and noisy variants of a large-scale urban taxi fare dataset, with a focus on comparative analysis across Graph Attention Networks (GAT), XGBoost, and TimesNet models.

\subsection{Data Pre-processing Data Handling}

The raw dataset consists of over 55 million rows of NYC taxi fare records. To enable efficient processing, chunk-wise loading and pre-processing were carried out using PySpark.

\textbf{Missing Value Imputation:} Missing values were identified in spatial attributes, including pickup and drop-off coordinates. These were imputed using the K-Nearest Neighbors (KNN) algorithm, which is effective in preserving geospatial patterns that null values may otherwise skew.

\textbf{1. Noise Injection:} To evaluate model robustness, Gaussian noise was artificially introduced across selected numerical features, including fare amount, passenger count, and location attributes. This emulates perturbations commonly observed in real-world datasets, which can be attributed to sensor errors, system noise, or transmission artifacts.

\textbf{2. Kolmogorov–Smirnov Test:} To verify whether noise altered data distributions significantly, the KS test was applied to the original and noise-perturbed columns. A p-value threshold of 0.05 was used to confirm statistically significant deviations.

\textbf{3. Feature Engineering:} Additional temporal features were derived from timestamps, including hour of the day, day of the week, and month. A Haversine formula was used to calculate the spatial distance between the pickup and drop-off coordinates, capturing trip displacement.

\textbf{Outlier Removal:} Outliers were detected and removed using an Interquartile Range (IQR)-based filtering method. Spatial and fare-related anomalies outside the expected bounds for NYC were excluded to prevent skewed learning.

\textbf{4. Autoencoder Denoising:} A deep autoencoder was trained on noisy columns to learn latent representations and reconstruct clean feature vectors. Denoised outputs were subsequently used in parallel with noisy versions for model training and evaluation.

\subsection{Graph Neural Network – GAT Model}

\textbf{1. Graph Construction:} Each taxi trip was treated as a node in a graph. Edges were constructed based on fare similarity and temporal adjacency between trips, thereby allowing the GAT model to learn from local contextual structures \cite{ref7}.

\textbf{2. GAT Architecture:} A multi-layer Graph Attention Network (GAT) was implemented for fare prediction. Attention coefficients were dynamically computed to assign importance to neighboring nodes, allowing for a selective focus on informative patterns during training \cite{ref3}.

\textbf{3. Training Scenarios:}
\begin{itemize}
    \item \textit{Clean Dataset:} Trained on original noise-free data.
    \item \textit{Noisy Dataset:} Trained on the noisy version to check robustness.
\end{itemize}

\textbf{4. Evaluation Metrics:} The model's performance was validated using the Mean Absolute Error (MAE), Mean Squared Error (MSE), and R² score.

\textbf{5. Uncertainty Estimation:} An ensemble of GAT models was used to generate uncertainty estimates through prediction variance. These estimates help quantify confidence in individual predictions.

\textbf{6. Reliability and OOD Testing:}
\begin{itemize}
    \item \textit{Calibration Plots:} Plots of predicted versus actual reliability were employed to determine prediction confidence \cite{ref6}.
    \item \textit{Out-of-Distribution (OOD) Testing:} GAT was evaluated on noise-perturbed samples to assess its robustness outside the training distribution \cite{ref5}.
\end{itemize}

\subsection{XGBoost Regression – Classical Baseline}

\textbf{1. Feature Matrix Preparation:} Denoised and preprocessed clean data were encoded and normalized. Spatial, temporal, and engineered features were incorporated \cite{ref9}.

\textbf{2. Model Training:} An XGBoost regressor was trained on both clean and noisy datasets using the squared error objective. An 80/20 train-test split was maintained for consistency in benchmarking \cite{ref10, ref11, ref12}.

\textbf{3. Hyperparameter Tuning:} A grid search strategy was employed to optimize the tree depth and L2 regularization (lambda) parameters. Early stopping was applied to prevent overfitting, with RMSE on validation sets used as the early stopping criterion.

\textbf{4. Assessment:} MAE, RMSE, R², and bespoke accuracy (percentage within ±10\% of actual fare) were calculated. Visual diagnostics included:
\begin{itemize}
    \item Scatter plots of predicted vs. actual fare
    \item Bin-wise MAE distribution
    \item Calibration curves
\end{itemize}

\subsection{TimesNet – Temporal Sequence Model}

\textbf{1. Preparation:}  To model temporal dependencies in taxi fare patterns, the dataset was segmented into sliding windows of size 10 over normalized features. These input sequences were used to predict the corresponding fare for each trip, enabling the model to learn from sequential dependencies within time-series patterns \cite{ref8}.

\textbf{2. Architecture Overview:}
\begin{itemize}
    \item \textit{Embedding Layer:} Maps input features to an embedding space \cite{ref13}.
    \item \textit{TimesBlocks:} 2D-Temporal Variation Blocks capture local and global time-series trends using reshaped convolutions.
    \item \textit{Regressor Head:} The final linear layer predicts a scalar fare.
\end{itemize}

\textbf{3. Training Configuration:} Optimizer: AdamW (learning rate = 5e-4, weight decay = 1e-5)

\textbf{Gradient Clipping:} max\_norm=1.0 \quad
\textbf{Scheduler:} ReduceLROnPlateau \quad
\textbf{Dropout:} 0.3 in each TimesBlock

\textbf{Early stopping:} patience=7 on validation MA

\textbf{Evaluation and Analysis:} Model performance was evaluated using the Mean Absolute Error (MAE), Root Mean Squared Error (RMSE), R-squared score, and custom accuracy, defined as the percentage of predictions within ±10\% of the actual fare. Visualization tools, such as scatter plots of predicted versus actual values, bin-wise MAE histograms, and calibration reliability curves, were used for model diagnostics \cite{ref14}.

\section{Dataset Overview}

The dataset utilized in this study was obtained from the publicly available NYC Yellow Taxi Trip Records, maintained by the New York City Taxi and Limousine Commission (TLC). Data was accessed via the NYC Open Data portal and includes over 55 million trip records.

Each row corresponds to a completed taxi trip and contains a mixture of geospatial, temporal, and financial features relevant to fare estimation.
Key features are:

pick\_up\_date-time:  Timestamp  of  ride  
pickup pickup\_longitude, pick\_up\_latitude: pickup location GPS coordinates.
drop\_off\_longitude, drop\_off\_latitude: Drop-off GPS coordinates
passenger\_count: Number of passengers fare\_amount: The fare paid by the passenger.

\begin{figure}[htbp]
  \centering
  \includegraphics[width=1.00\linewidth]{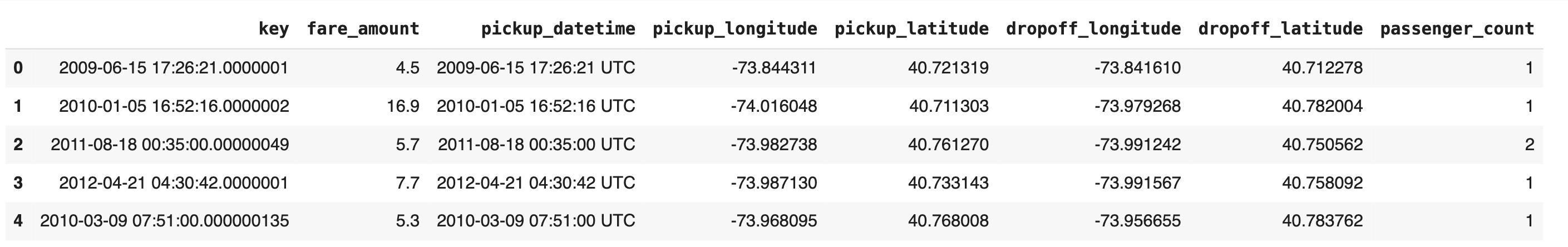} 
  \caption{Sample Dataset visualization.}
  \label{fig:sample_dataset}
\end{figure}

These features are crucial as they encapsulate key elements such as distance traveled, time-of-day effects, passenger behavior, and urban spatial dynamics.~\autoref{fig:sample_dataset} shows the Sample Dataset.

Due to the large volume of records and diverse feature types, significant pre-processing and distributed computation techniques were necessary. Chunk-wise processing using PySpark and Dask enabled scalable handling of the data.

Several key challenges were encountered:

\textbf{Geospatial Complexity:} Accurate fare estimation requires modeling spatial dependencies between pickup and drop-off points across the urban grid.

\textbf{Temporal Variability:} Fare prices vary by hour, weekday, and season, reflecting fluctuations during rush hours, holidays, and off-peak periods.

\textbf{Data Quality Issues:} Frequent missing entries, outliers, and synthetic noise in numerical columns necessitate robust pre-processing steps, including KNN imputation, IQR-based filtering, and deep learning-based denoising.
The dataset renders the problem a complex, high-dimensional regression task under real-world conditions, making it suitable for evaluating model robustness and generalizability.

\begin{figure*}[ht]
    \centering
    \includegraphics[width= 0.7\linewidth]{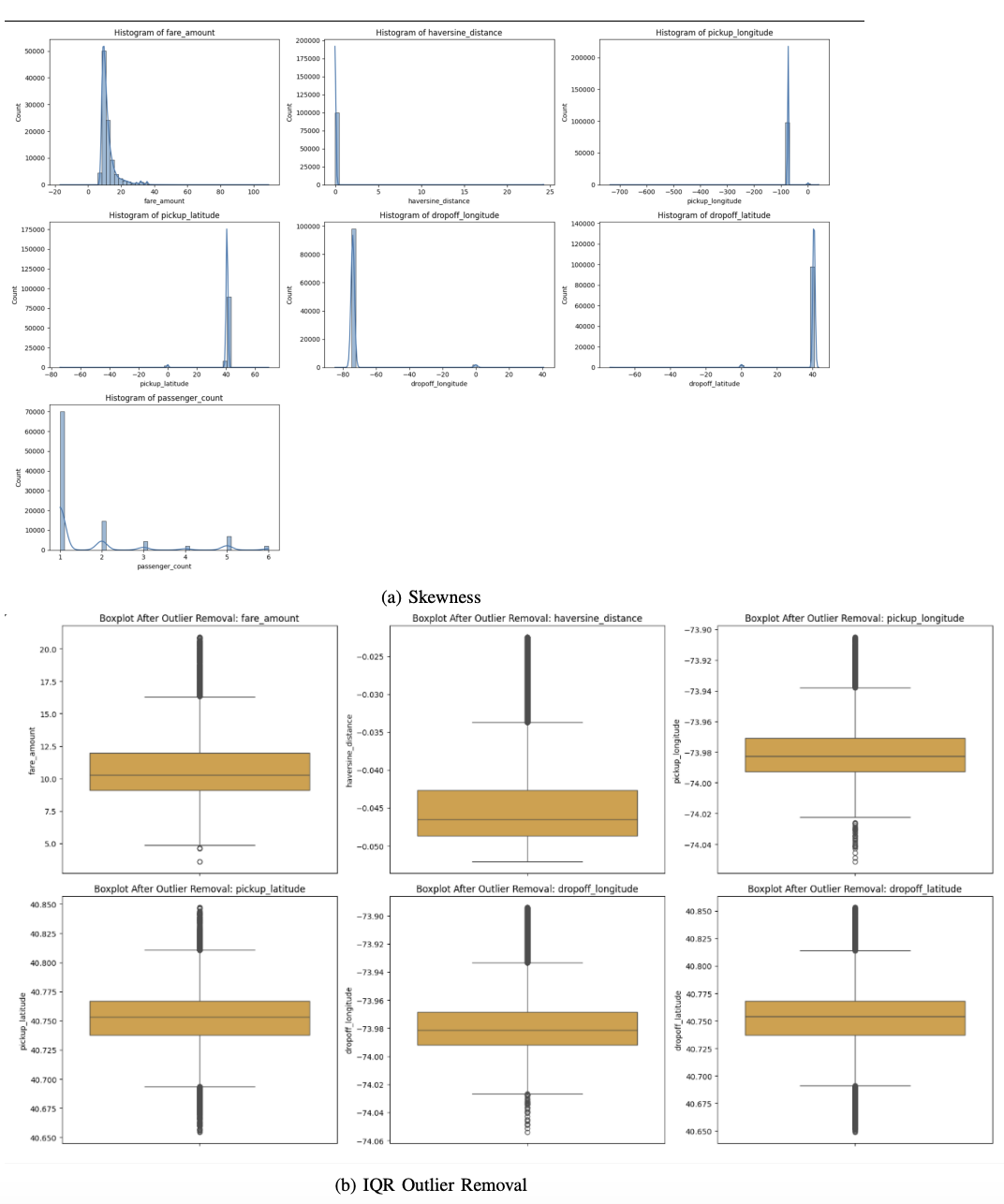}  
    \vspace{1mm}
    \caption{(a) Skewness and (b) IQR Outlier Removal from the dataset.}
    \label{fig:skew_iqr}
\end{figure*}
\section{Results and Analysis}

\subsection{\textit{Gaussian Noise Injection Effect}}

\textbf{1. Objective:}

This section examines the impact of Gaussian noise injection on the statistical properties of key features and the performance of the downstream model.

\textbf{2. Observations:}

Gaussian noise was applied to spatial and numeric columns, simulating real-world inconsistencies such as GPS drift or system-level inaccuracies. The noise injection introduced measurable changes in feature distributions.~\autoref{fig:skew_iqr} shows the skewness before and after perturbation, revealing altered distribution symmetry and presents outlier removal using inter-quartile range (IQR) filtering, showing the efficacy of denoising. \cite{ref7} \cite{ref1}.

~\autoref{fig:geo_fare_features} below shows the impact of Gaussian noise on each column to check which is more robust to noise.

\begin{figure*}[ht]
\centering

\begin{minipage}[t]{0.48\textwidth}
  \centering
  \includegraphics[width=0.95\linewidth]{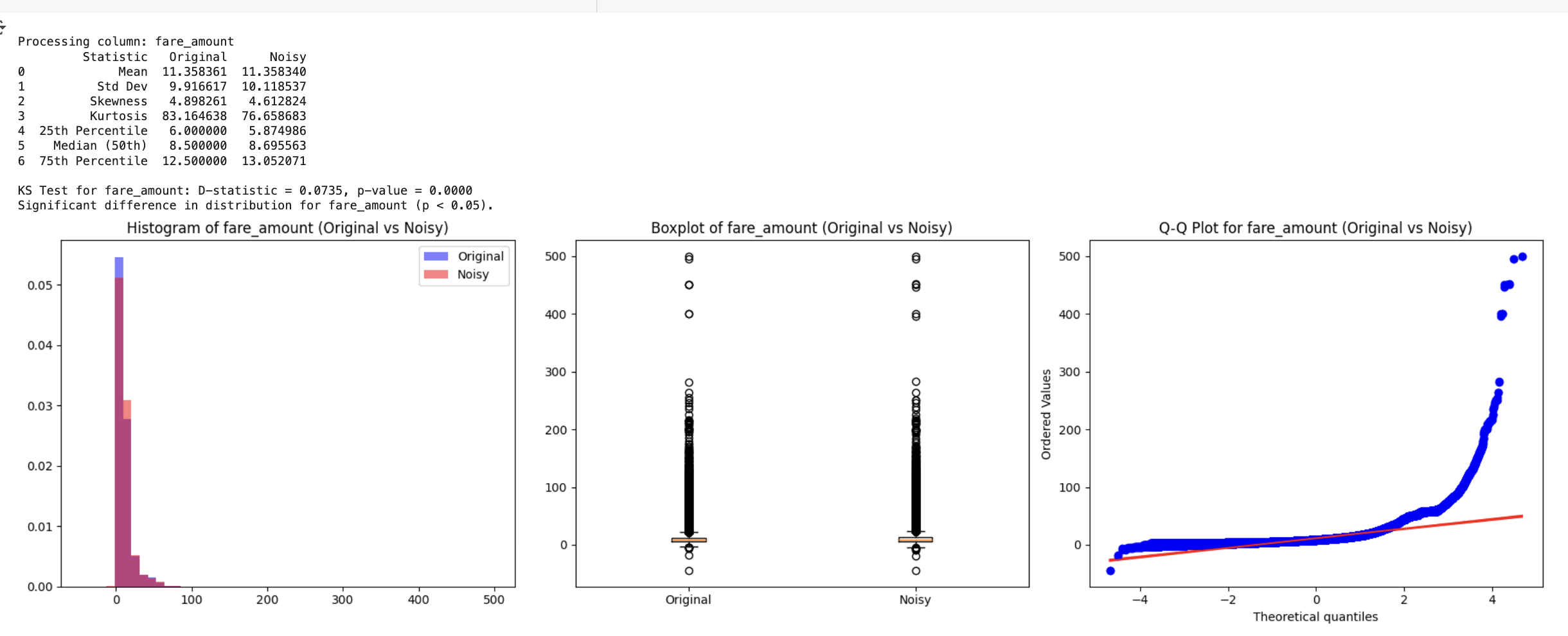}
  \caption*{(a) Fare Amount}
\end{minipage}
\hfill
\begin{minipage}[t]{0.48\textwidth}
  \centering
  \includegraphics[width=0.95\linewidth]{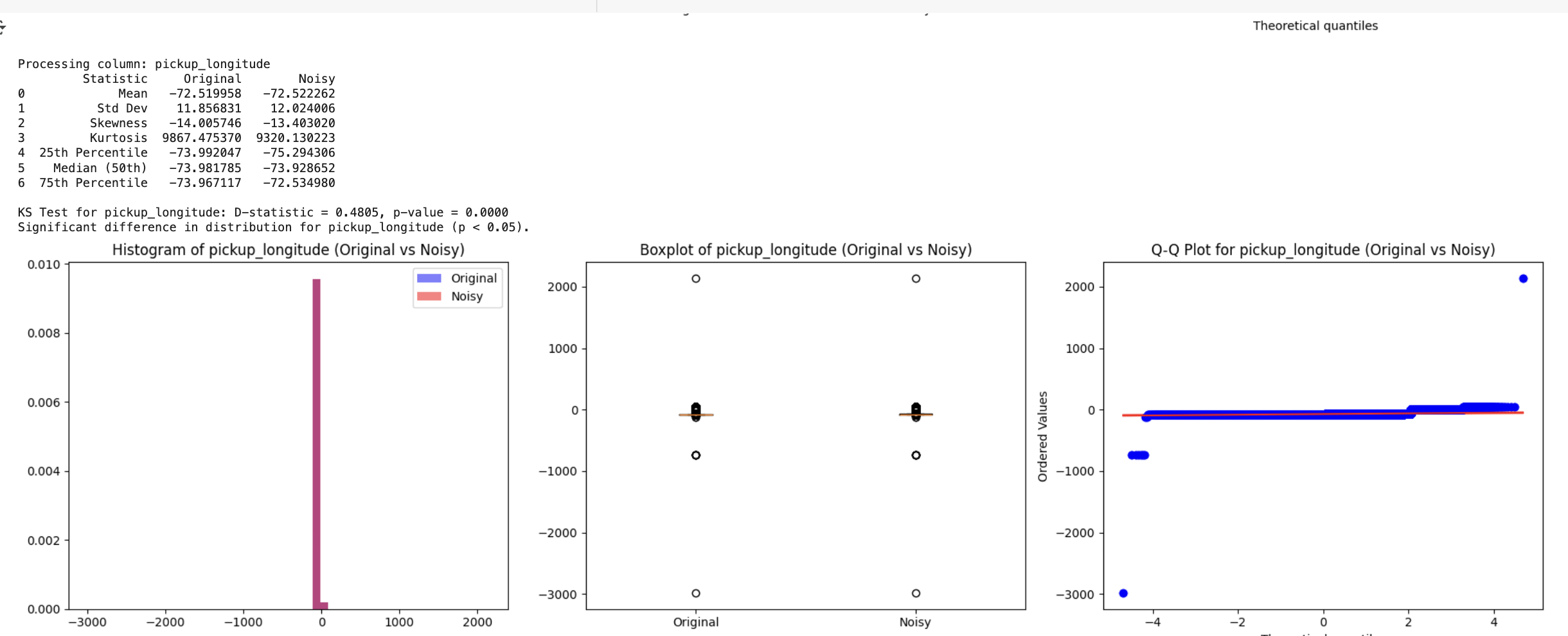}
  \caption*{(b) Pickup Longitude}
\end{minipage}

\vspace{5mm}

\begin{minipage}[t]{0.48\textwidth}
  \centering
  \includegraphics[width=0.95\linewidth]{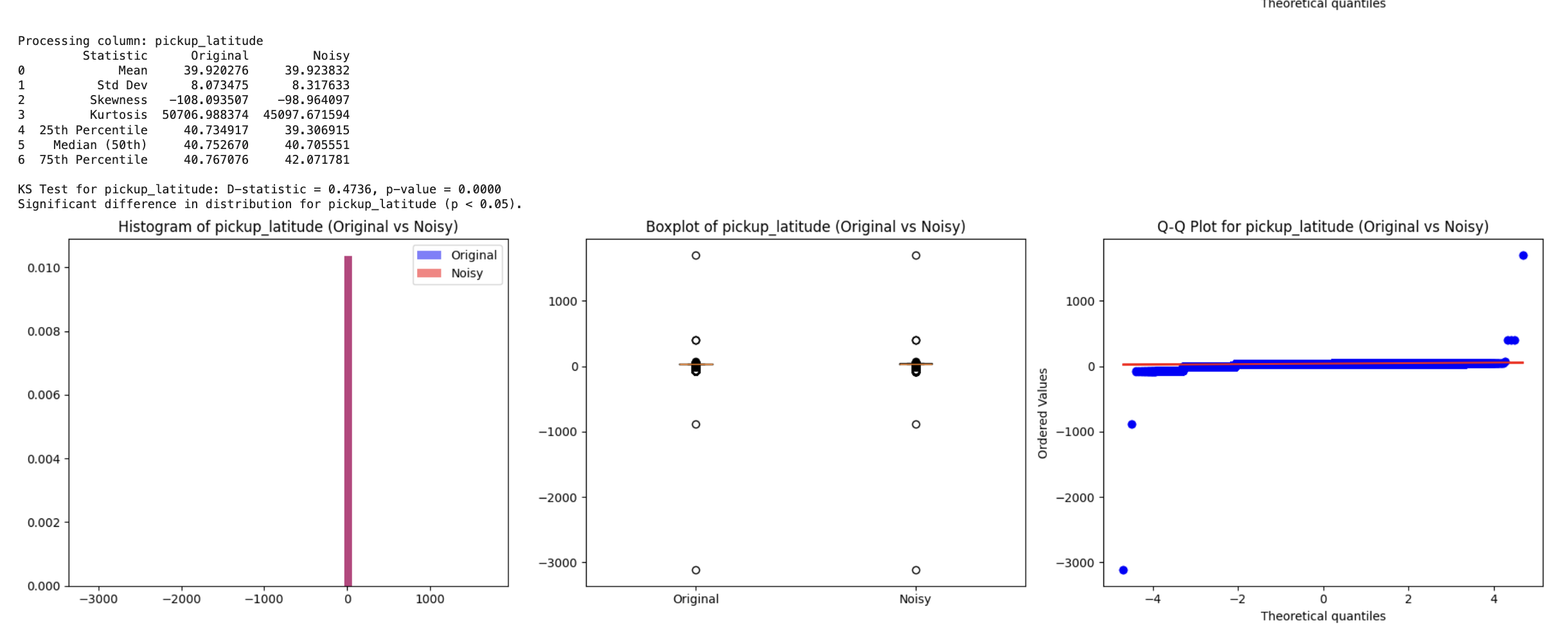}
  \caption*{(c) Pickup Latitude}
\end{minipage}
\hfill
\begin{minipage}[t]{0.48\textwidth}
  \centering
  \includegraphics[width=0.95\linewidth]{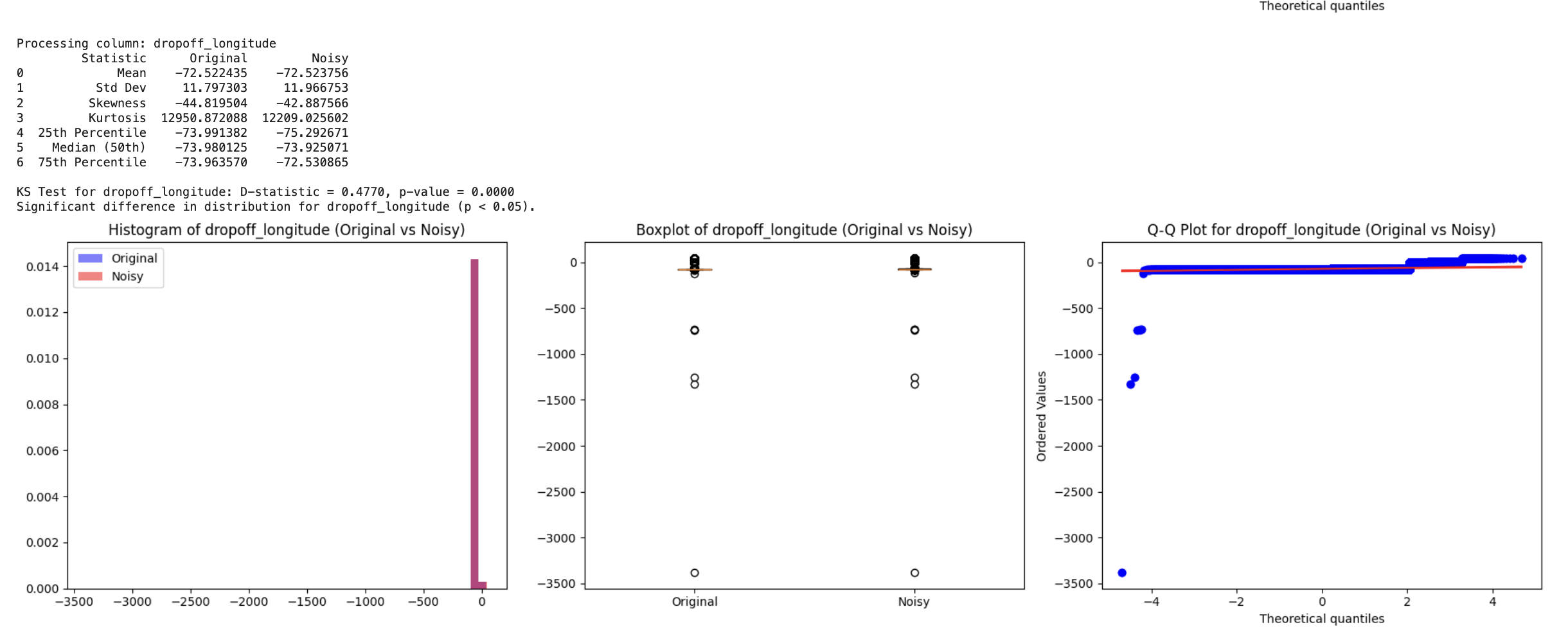}
  \caption*{(d) Drop-off Longitude}
\end{minipage}

\vspace{5mm}

\begin{minipage}[t]{0.48\textwidth}
  \centering
  \includegraphics[width=0.95\linewidth]{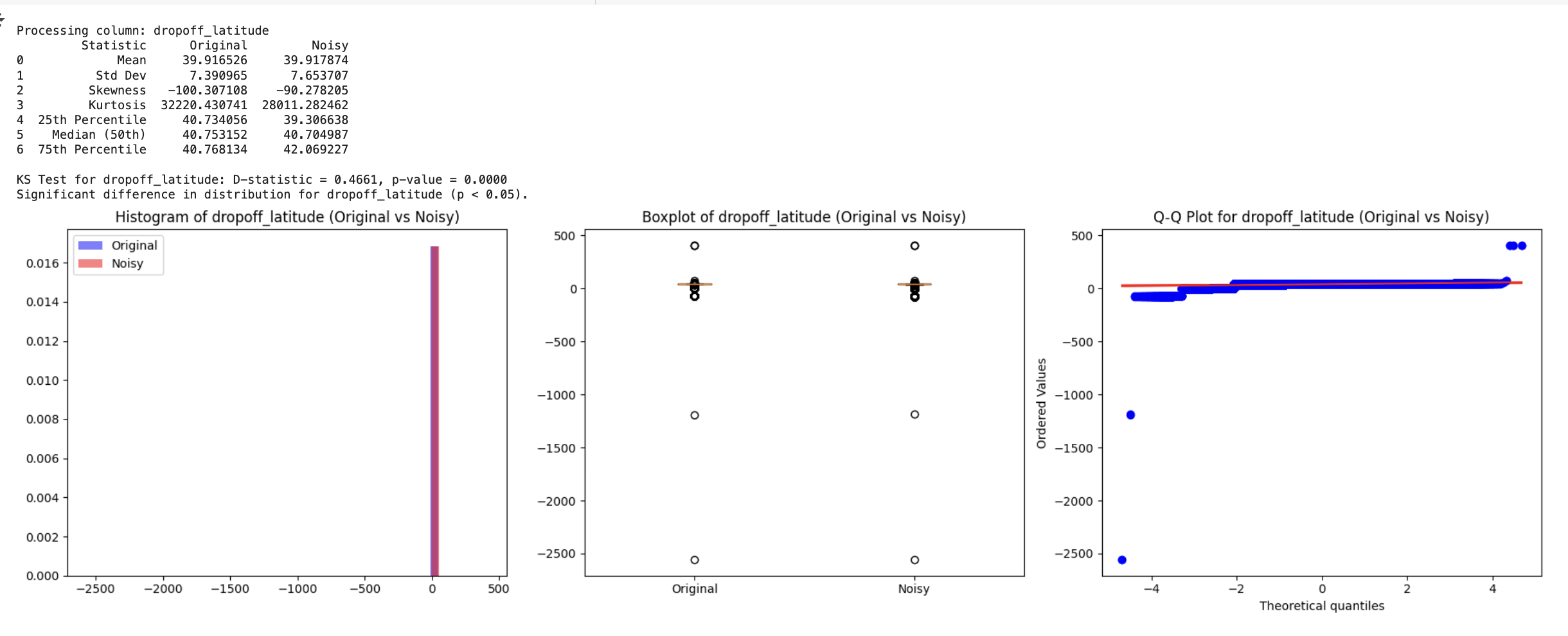}
  \caption*{(e) Drop-off Latitude}
\end{minipage}

\caption{Visualizations of key geospatial and fare-related features.}
\label{fig:geo_fare_features}
\end{figure*}

\subsection{\textit{GAT Model Performance:}}

The impact of noise on the GAT model's regression metrics is shown in~\autoref{tab:gat_performance}. The performance on clean data suggests that GAT captures spatial dependencies well under ideal conditions. However, when Gaussian noise is introduced, a significant degradation in model performance is observed across all metrics.

\begin{table}[H]
\centering
\renewcommand{\arraystretch}{1.2} 
\begin{tabular}{|l|c|c|}
\hline
\textbf{Metric} & \textbf{Clean Data} & \textbf{Noisy Data} \\
\hline
MAE & 1.1057 & 2.5036 \\
MSE & 2.4414 & 11.4375 \\
$R^2$ Score & 0.2153 & 1.1237 \\
\hline
\end{tabular}
\vspace{2mm}
\caption{Performance of GAT model on clean vs. noisy datasets}
\label{tab:gat_performance}
\end{table}

\textbf{1. Key Observations:}

\textbf{Calibration:} The model demonstrates near-perfect calibration on the clean dataset, while noise severely disrupts its ability to produce reliable predictions. This indicates a strong sensitivity to data perturbations.

\textbf{Prediction Uncertainty:} On clean inputs, GAT maintains narrow uncertainty bands and consistent confidence scores. However, noisy data introduces higher predictive variance, suggesting reduced model trustworthiness under perturbation \cite{ref6}.

\textbf{Bin-wise MAE:} Clean data yields consistent MAE across bins, whereas noisy inputs lead to elevated errors, particularly in higher-fare brackets. This indicates that GAT becomes increasingly unreliable for extreme cases under noisy conditions \cite{ref6}.

\textbf{Reliability:} The model's reliability degrades substantially under noisy inputs, with clean data exhibiting stable performance and perturbed inputs resulting in erratic prediction behavior.

\vspace{1mm}
\textbf{2. Out-of-Distribution (OOD) Testing (GAT):}
The out-of-distribution performance, as shown in~\autoref{tab:gat_ood}, reveals a notable increase in prediction error when tested on noise-injected data. Although MAE and MSE remain comparable to those seen in~\autoref{tab:gat_performance}, the R² score drops into an unfavorable regime, indicating an ineffective capture of variance.

\begin{table}[H]
\centering
\begin{tabular}{|l|c|c|}
\hline
\textbf{Metric} & \textbf{Clean Data} & \textbf{Noisy Data} \\
\hline
MAE & 1.1475 & 2.5036 \\
MSE & 2.4414 & 11.4374 \\
R\textsuperscript{2} & 0.2153 & -1.1237 \\
\hline
\end{tabular}
\vspace{1mm} 
\caption{GAT OOD}
\label{tab:gat_ood}
\end{table}

\begin{figure*}[!t]
  \centering
  \begin{minipage}{0.32\textwidth}
    \centering
    \includegraphics[width=1.1\linewidth]{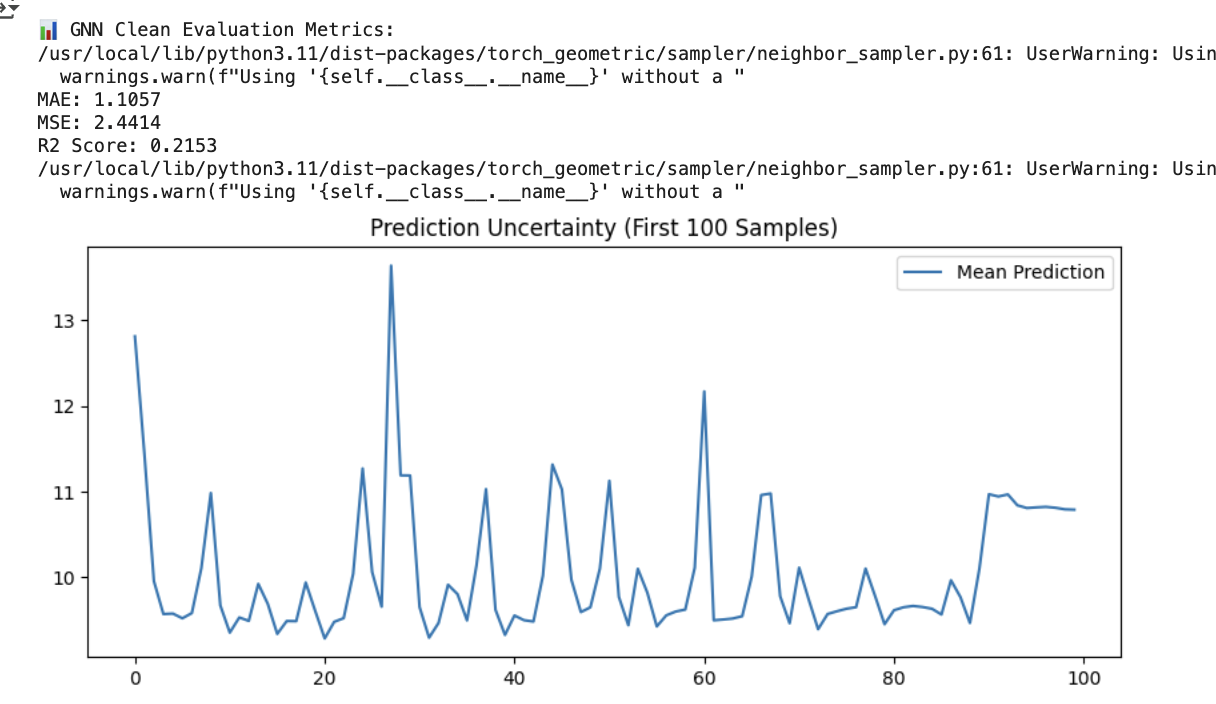}
    \caption*{(a) GAT Uncertainty}
  \end{minipage}
  \hfill
  \begin{minipage}{0.32\textwidth}
    \centering
    \includegraphics[width=1.1\linewidth]{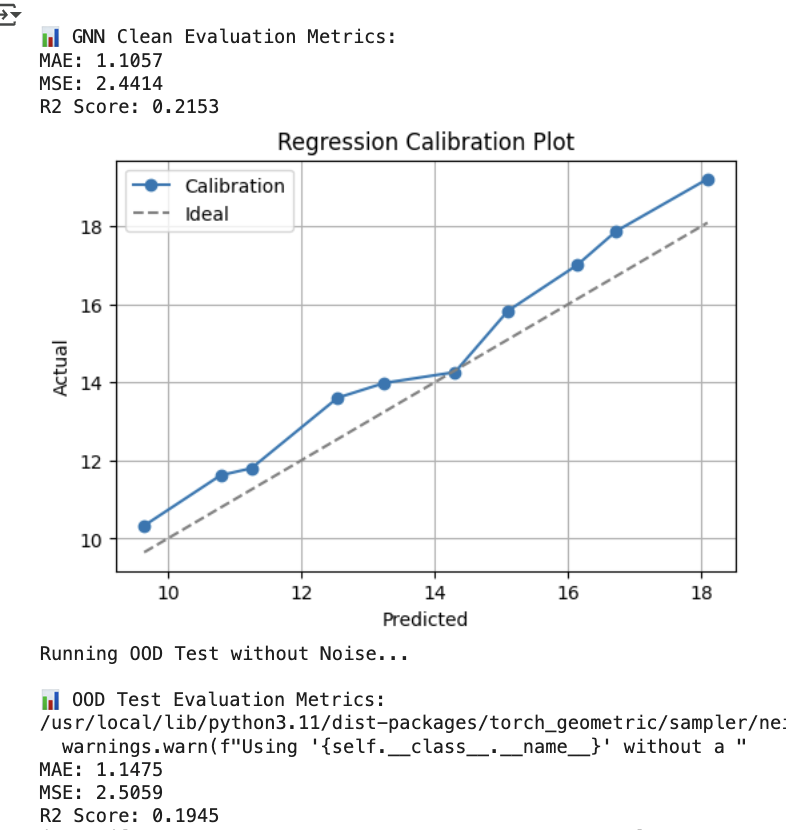}
    \caption*{(b) GAT Calibration}
  \end{minipage}
  \hfill
  \begin{minipage}{0.32\textwidth}
    \centering
    \includegraphics[width=1.1\linewidth]{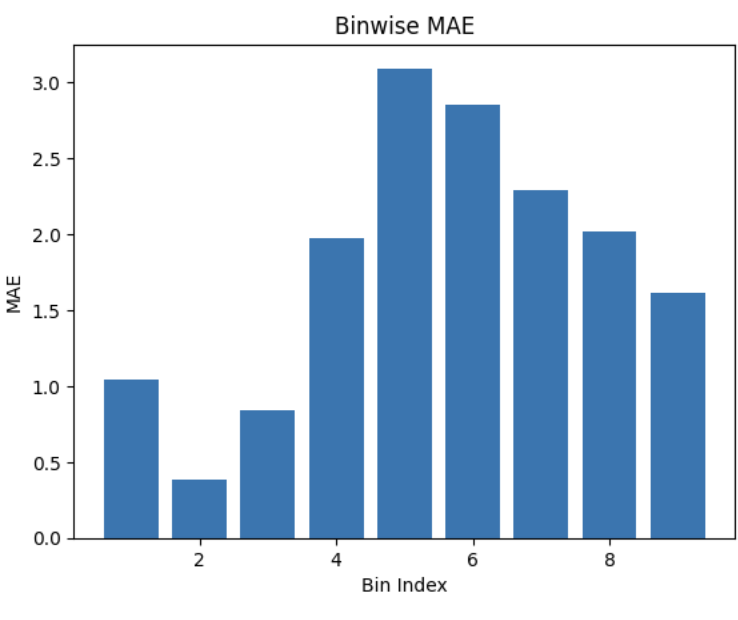}
    \caption*{(c) GAT Bin-wise MAE}
  \end{minipage}
  \vspace{-2mm}
  \caption{\textit{GAT Denoised Data}}
  \label{fig:gat_denoised}
\end{figure*}

\begin{figure*}[!t]
  \centering
  \begin{minipage}{0.32\textwidth}
    \centering
    \includegraphics[width=1.1\linewidth]{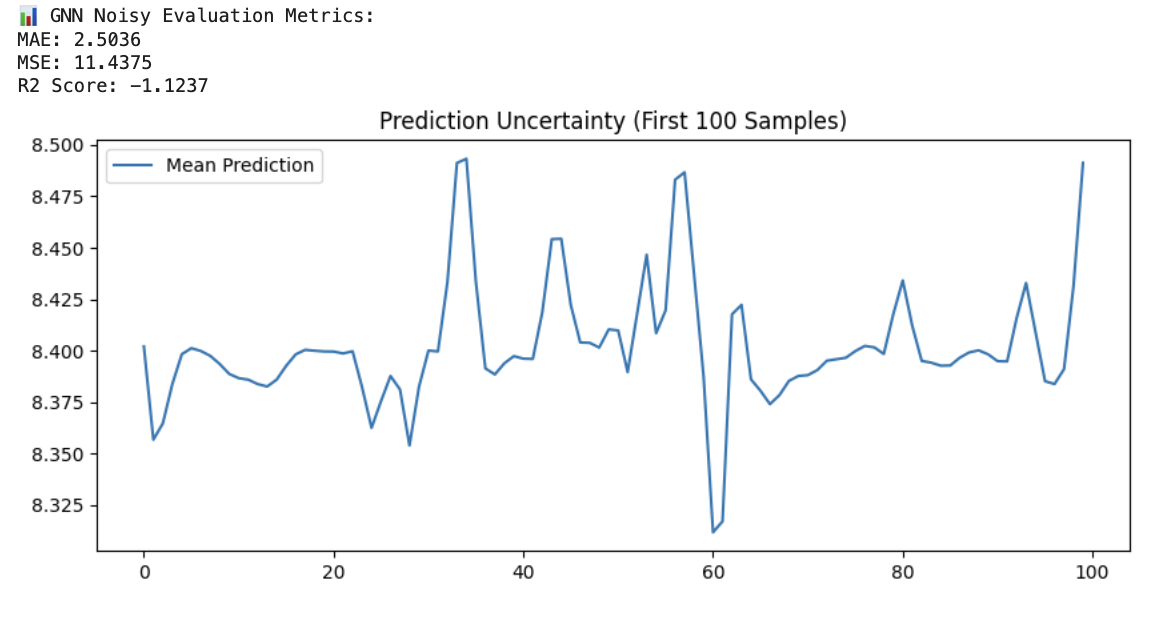}
    \caption*{(a) GAT Uncertainty}
  \end{minipage}
  \hfill
  \begin{minipage}{0.32\textwidth}
    \centering
    \includegraphics[width=1.1\linewidth]{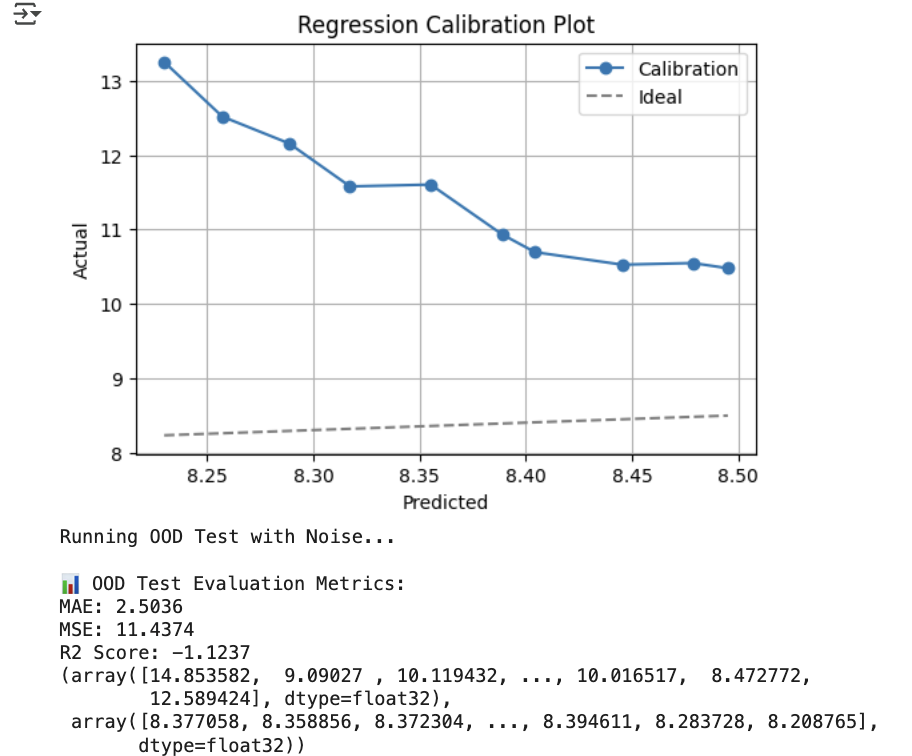}
    \caption*{(b) GAT Calibration}
  \end{minipage}
  \hfill
  \begin{minipage}{0.32\textwidth}
    \centering
    \includegraphics[width=1.1\linewidth]{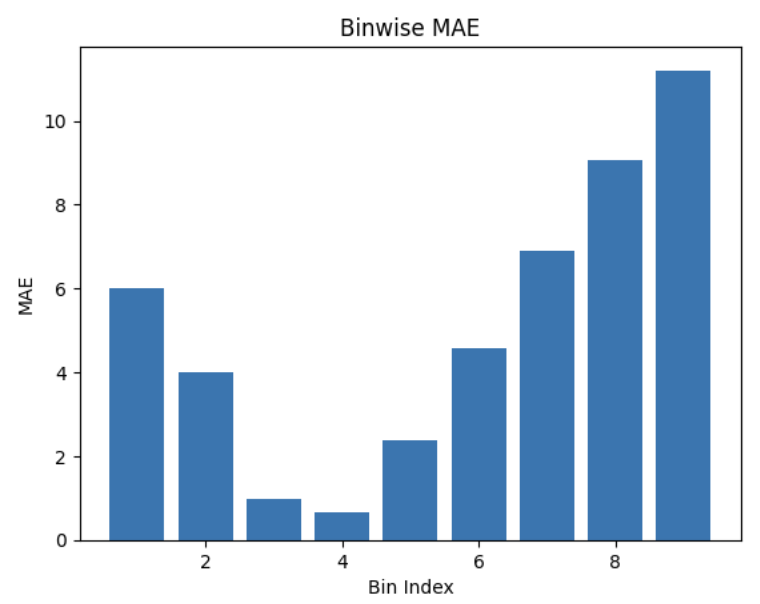}
    \caption*{(c) GAT Bin-wise MAE}
  \end{minipage}
  \vspace{-2mm}
  \caption{\textit{GAT Noisy Data}}
  \label{fig:gat_noised}
\end{figure*}

\textbf{3. Conclusion:} The \textbf{GAT} model demonstrates strong capabilities when trained and evaluated on clean datasets. However, its robustness is compromised under noisy settings, manifesting in poor generalization, degraded calibration, and unreliable uncertainty estimation. These results indicate the need for incorporating noise-aware regularization or architectural enhancements to improve robustness.~\autoref{fig:gat_denoised} \&~\autoref{fig:gat_noised} show the performance of GAT with noise and denoised data.

\subsection{\textit{XGBoost Model Performance}}

XGBoost's performance across clean and noisy datasets is summarized in~\autoref{tab:xg_performance}. The model exhibits high predictive accuracy and minimal degradation in performance under noisy inputs, illustrating its suitability for high-dimensional tabular tasks.

\begin{table}[H]
\centering
\begin{tabular}{|l|c|c|}
\hline
\textbf{Metric} & \textbf{Clean Data} & \textbf{Noisy Data} \\
\hline
MAE & 0.1040 & 0.8203 \\
MSE & 0.0279 & 1.4407 \\
R\textsuperscript{2} & 0.9910 & 0.7326 \\
\hline
\end{tabular}
\vspace{1mm}
\caption{XGBoost Performance}
\label{tab:xg_performance}
\end{table}
\vspace{-5mm}
\textbf{1. Key Observations:}

\textbf{Calibration:} Calibration plots reveal near-ideal alignment between predicted and actual fares for clean data. Although minor deterioration is observed under noise, the model retains a well-calibrated output distribution.

\textbf{Prediction Uncertainty:} The model provides tight confidence intervals under clean conditions. Under noise, uncertainty bands expand slightly but remain stable relative to GAT, highlighting better resilience \cite{ref6}.

\textbf{Bin-wise MAE:} XGBoost maintains a flat error profile across fare bins, with limited error spikes at extreme values. Its stability across low- and high-fare segments confirms its reliability \cite{ref6}.

\textbf{Reliability:} Prediction reliability is consistently high regardless of noise, showcasing the model's robustness and generalization capabilities.

\vspace{2mm}
\textbf{2. Out-of-Distribution (OOD) Test (XGBoost):}
As shown in~\autoref{tab:xg_ood}, XGBoost outperforms GAT in OOD settings. While error metrics increase slightly on the noise-perturbed dataset, the drop in R2 is minimal. This stability reflects XGBoost's ability to generalize well across variations in data distribution.

\begin{table}[H]
\centering
\begin{tabular}{|l|c|c|}
\hline
\textbf{Metric} & \textbf{Clean Data} & \textbf{Noisy Data} \\
\hline
MAE & 0.5921 & 0.8504 \\
MSE & 0.7006 & 1.5116 \\
R\textsuperscript{2} & 0.7748 & 0.7195 \\
\hline
\end{tabular}
\vspace{1mm}
\caption{OOD XGBoost}
\label{tab:xg_ood}
\end{table}

\begin{figure*}[!t]
  \centering
  \begin{minipage}{0.3\textwidth}
    \centering
    \includegraphics[width=1.1\linewidth]{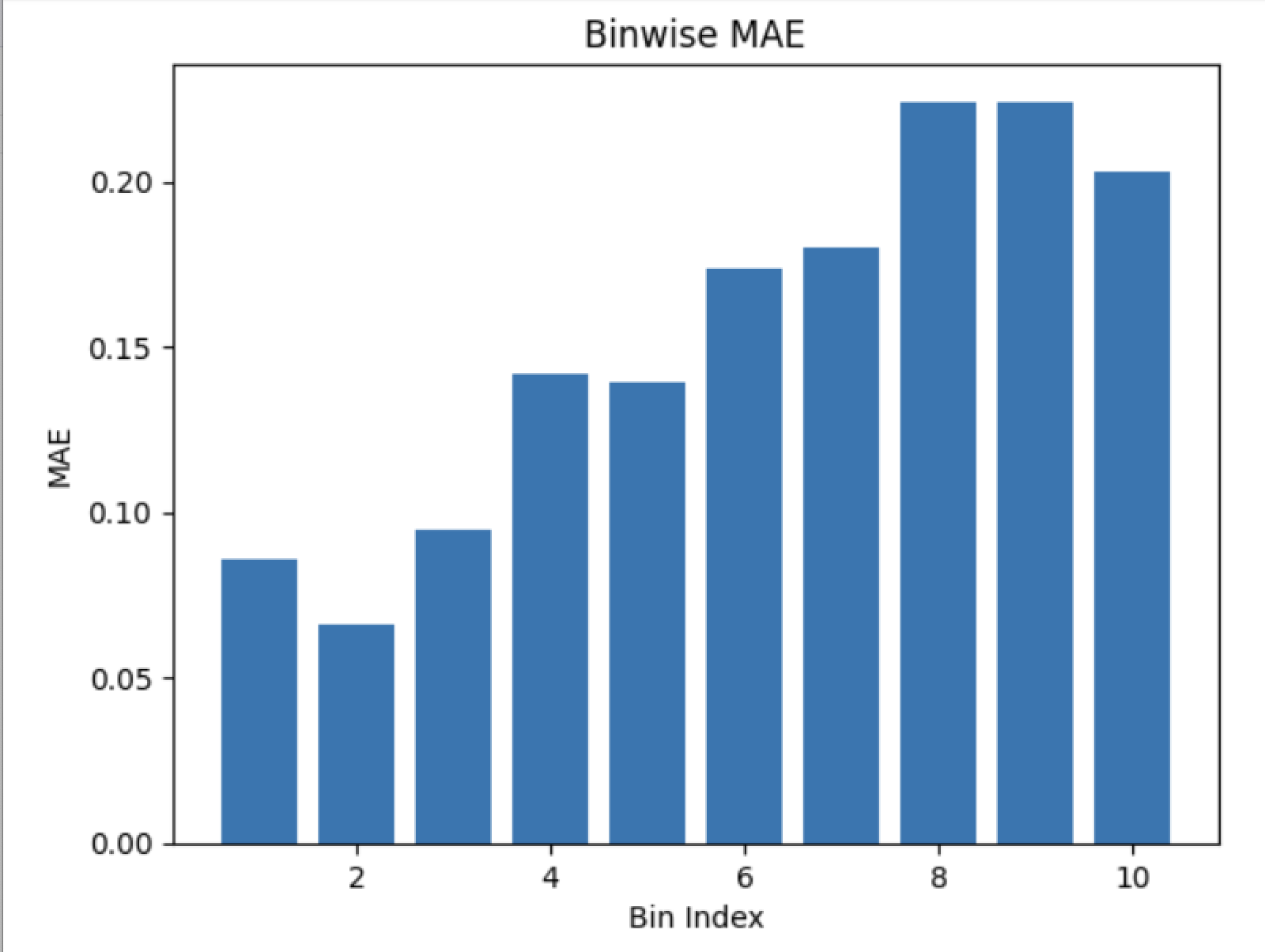}
    \caption*{(a) XGBoost Bin-wise MAE}
  \end{minipage}
  \hfill
  \begin{minipage}{0.3\textwidth}
    \centering
    \includegraphics[width=1.1\linewidth]{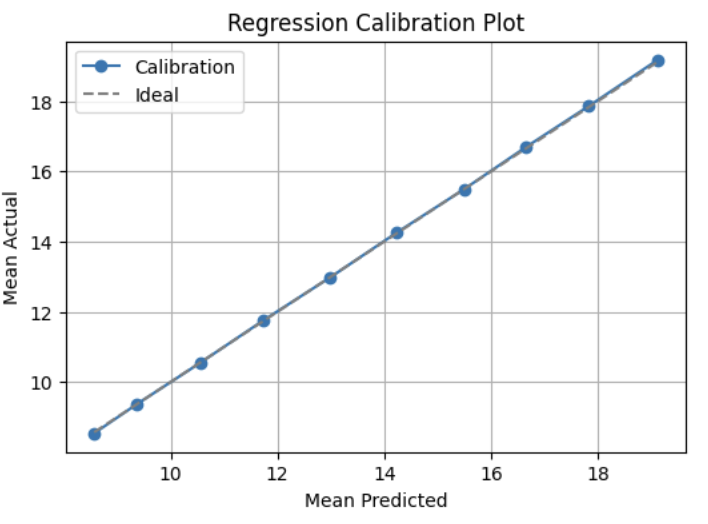}
    \caption*{(b) XGBoost Calibration}
  \end{minipage}
  \hfill
  \begin{minipage}{0.3\textwidth}
    \centering
    \includegraphics[width=1.1\linewidth]{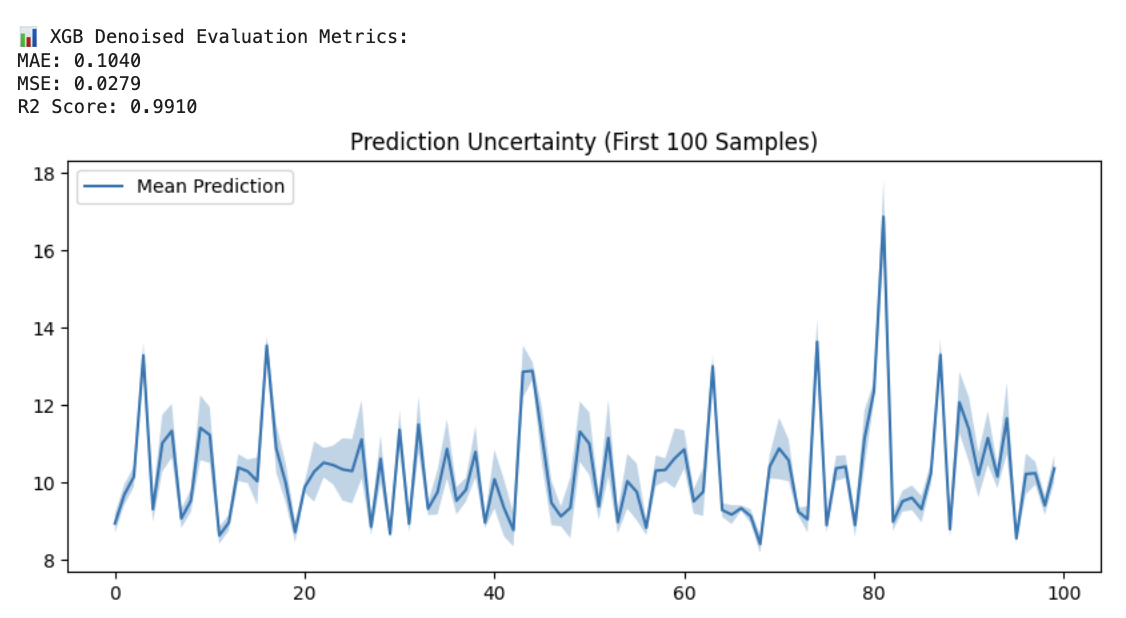}
    \caption*{(c) XGBoost Uncertainty}
  \end{minipage}
  \vspace{-2mm}
  \caption{\textit{XGBoost Denoised Data}}
  \label{fig:xg_denoised}
\end{figure*}

\begin{figure*}[!t]
  \centering
  \begin{minipage}{0.3\textwidth}
    \centering
    \includegraphics[width=1.1\linewidth]{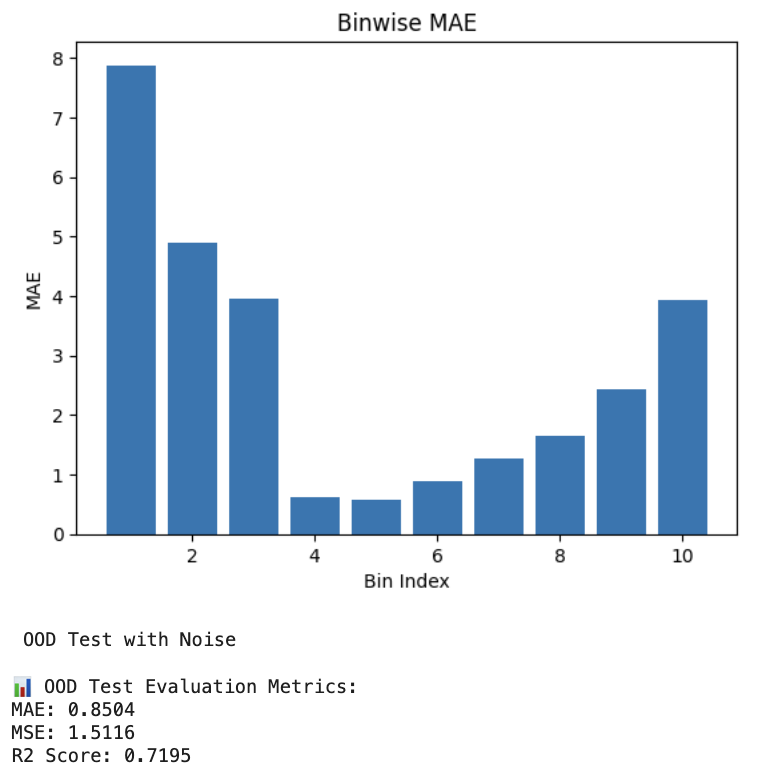}
    \caption*{(a) XGBoost Bin-wise MAE}
  \end{minipage}
  \hfill
  \begin{minipage}{0.3\textwidth}
    \centering
    \includegraphics[width=1.1\linewidth]{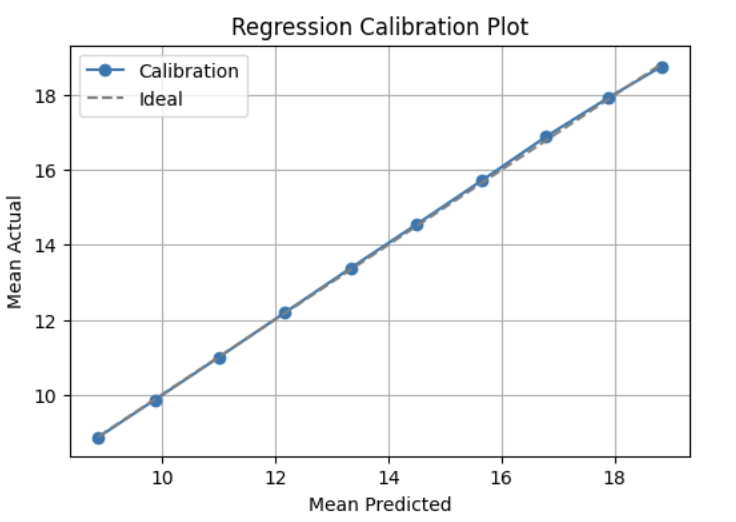}
    \caption*{(b) XGBoost Calibration}
  \end{minipage}
  \hfill
  \begin{minipage}{0.3\textwidth}
    \centering
    \includegraphics[width=1.1\linewidth]{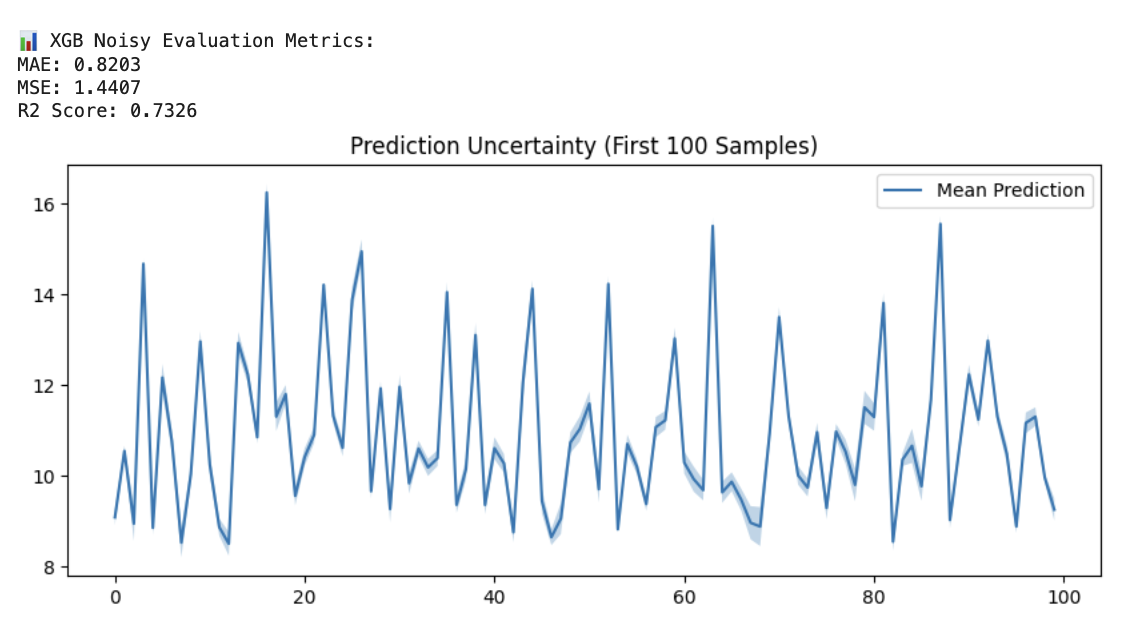}
    \caption*{(c) XGBoost Uncertainty}
  \end{minipage}
  \vspace{-2mm}
  \caption{\textit{XGBoost Noisy Data}}
  \label{fig:xg_noised}
\end{figure*}

\textbf{3. Conclusion:} The \textbf{XGBoost} demonstrates superior performance under both clean and noisy conditions. Its strong calibration, consistent uncertainty estimates, and generalization under OOD conditions validate its applicability for real-world fare prediction tasks. In contrast to GAT and TimesNet, XGBoost emerges as a reliable baseline for both structured tabular learning and robustness under data perturbation.~\autoref{fig:xg_denoised} \&~\autoref{fig:xg_noised} show the performance of Classic XGBoost with noisy and denoised data.

\subsection{\textit{TimesNet Model Performance}}
The performance metrics of the TimesNet model under clean and noisy data conditions are summarized in~\autoref{tab:timesnet_performance}. As a multivariate temporal deep learning architecture, TimesNet demonstrates moderate gains after denoising but remains challenged by noise-induced variability.

\begin{table}[H]
\centering
\begin{tabular}{|l|c|c|}
\hline
\textbf{Metric} & \textbf{Clean Data} & \textbf{Noisy Data} \\
\hline
MAE & 1.3549 & 1.7303 \\
MSE & 3.3534 & 5.8835 \\
R\textsuperscript{2} & -0.0779 & -0.0919 \\
\hline
\end{tabular}
\vspace{1mm}
\caption{TimesNet Performance}
\label{tab:timesnet_performance}
\end{table}

\textbf{1. Key Findings:}

\textbf{Calibration:} Slight improvements in calibration are observed in the clean dataset following denoising. However, the model fails to maintain ideal calibration, with outputs still deviating from the expected prediction-target alignment.

\textbf{Prediction Uncertainty:} On clean inputs, TimesNet exhibits relatively smooth and coherent profiles of predictive uncertainty. Under noisy conditions, however, the uncertainty becomes erratic and unreliable, indicating a loss of model confidence and robustness \cite{ref6}.

\textbf{Bin-wise MAE:} Error analysis across fare bins shows monotonic improvements for clean data. In contrast, the noise-injected variant produces a U-shaped profile, where error rates rise sharply for both low and high fare bins, underscoring the model's vulnerability to distributional extremes \cite{ref6}.

\textbf{Reliability:} Although reliability metrics improve after pre-processing, persistent inconsistencies across fare ranges indicate that the model still struggles to generalize effectively, particularly under perturbed input conditions.

\vspace{2mm}
\textbf{2. Out-of-Distribution (OOD) Testing (TimesNet):}
Results from the OOD evaluation are presented in~\autoref{tab:timesnet_ood}. The model's performance deteriorates under noise, with the R2 metric turning negative—indicative of predictive distributions failing to capture true variance. Although the denoised dataset provides marginally better performance, the model remains unstable.

\begin{figure*}[!t]
  \centering
  \begin{minipage}{0.3\textwidth}
    \centering
    \includegraphics[width=1.1\linewidth]{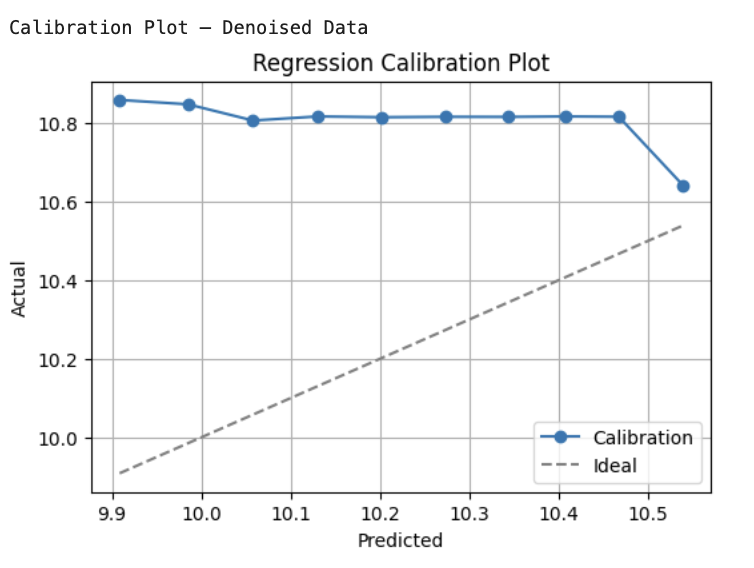}
    \caption*{(a) TimesNet Calibration}
  \end{minipage}
  \hfill
  \begin{minipage}{0.3\textwidth}
    \centering
    \includegraphics[width=1.1\linewidth]{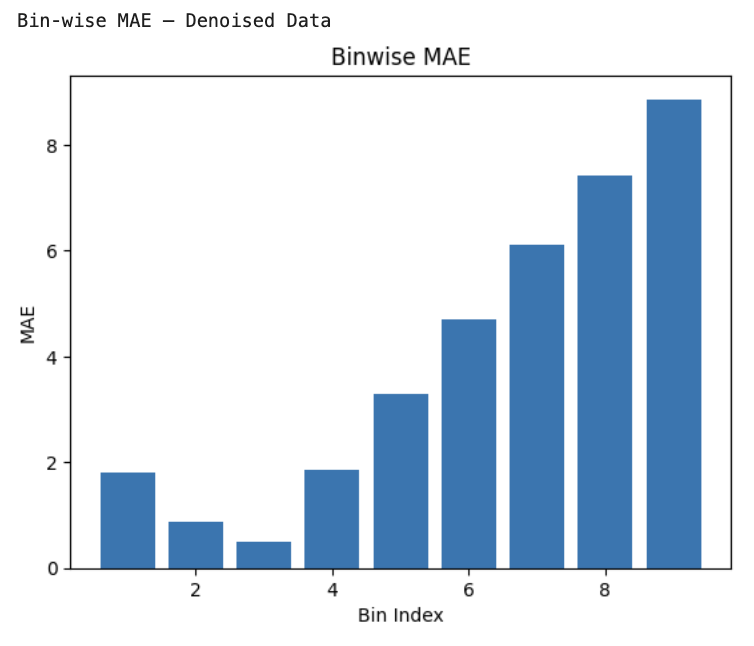}
    \caption*{(b) TimesNet Bin-wise MAE }
  \end{minipage}
  \hfill
  \begin{minipage}{0.3\textwidth}
    \centering
    \includegraphics[width=1.1\linewidth]{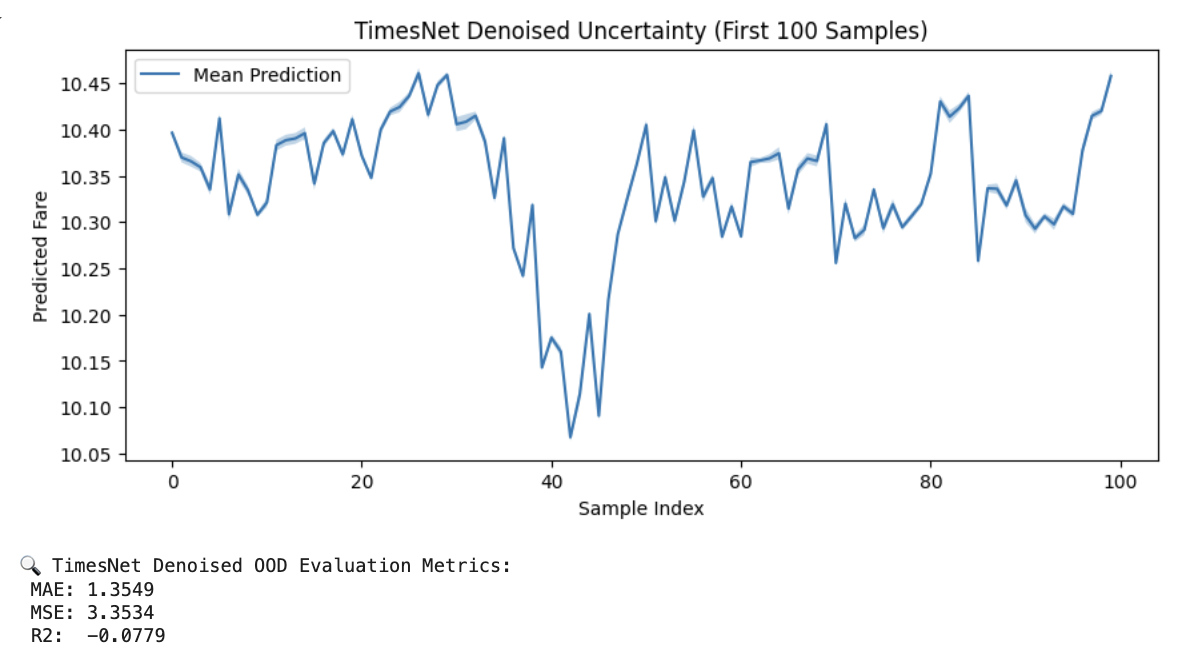}
    \caption*{(c) TimesNet Uncertainty}
  \end{minipage}
  \vspace{-2mm}
  \caption{\textit{TimesNet Denoised Data}}
  \label{fig:timesnet_denoised}
\end{figure*}

\begin{figure*}[!t]
  \centering
  \begin{minipage}{0.3\textwidth}
    \centering
    \includegraphics[width=1.1\linewidth]{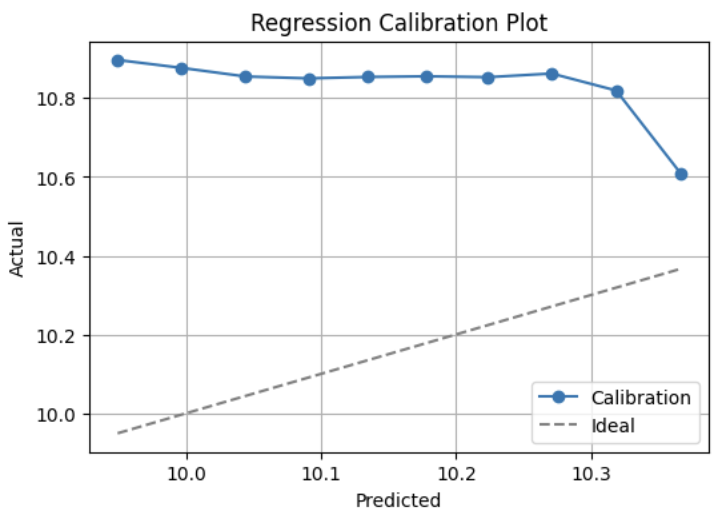}
    \caption*{(a) TimesNet Calibration}
  \end{minipage}
  \hfill
  \begin{minipage}{0.3\textwidth}
    \centering
    \includegraphics[width=1.1\linewidth]{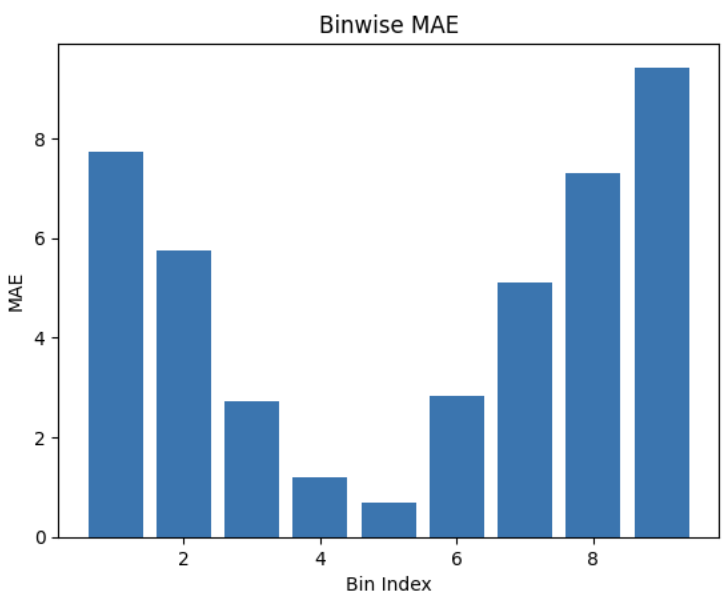}
    \caption*{(b) TimesNet Bin-wise MAE}
  \end{minipage}
  \hfill
  \begin{minipage}{0.3\textwidth}
    \centering
    \includegraphics[width=1.1\linewidth]{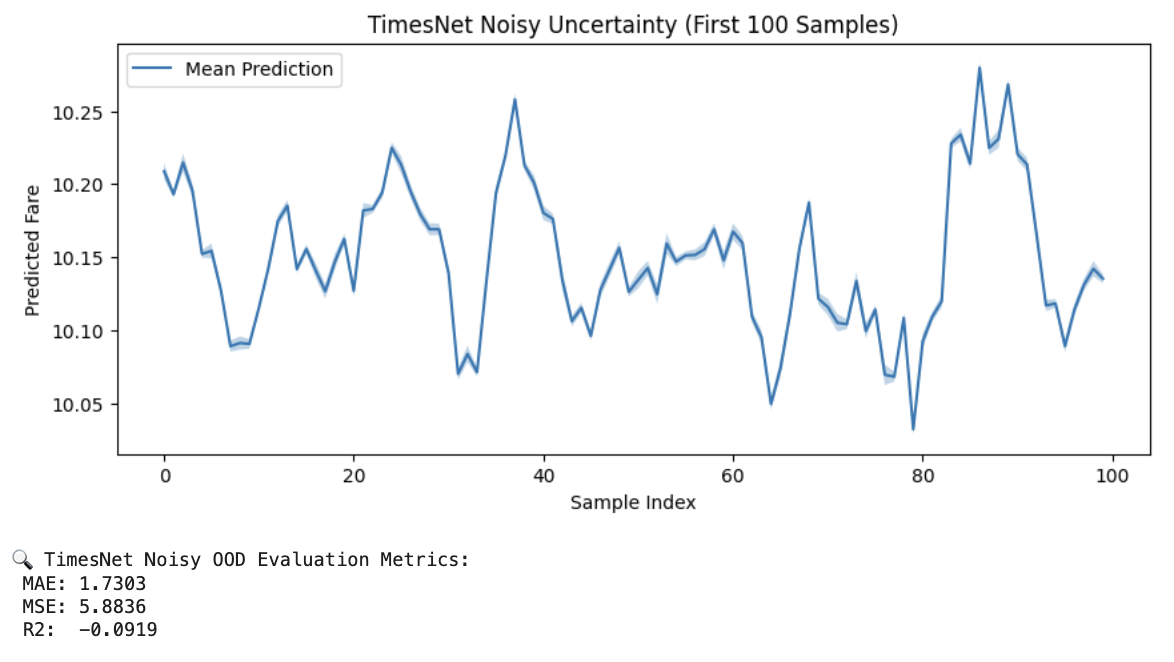}
    \caption*{(c) TimesNet Uncertainty}
  \end{minipage}
  \vspace{-2mm}
  \caption{\textit{TimesNet Noisy Data}}
  \label{fig:timesnet_noised}
\end{figure*}

\begin{table}[H]
\centering
\begin{tabular}{|l|c|c|}
\hline
\textbf{Metric} & \textbf{Clean Data} & \textbf{Noisy Data} \\
\hline
MAE & 1.3549 & 1.7303 \\
MSE & 3.3534 & 5.8835 \\
R\textsuperscript{2} & -0.0779 & -0.0919 \\
\hline
\end{tabular}
\vspace{1mm}
\caption{OOD TimesNet}
\label{tab:timesnet_ood}
\end{table}
\vspace{-4mm}
\textbf{3. Conclusion:} \textbf{TimesNet} demonstrates partial resilience to noise through data denoising, yielding modest improvements in calibration and uncertainty handling. Nonetheless, the model fails to sustain consistent performance under distributional shifts, with persistent degradation in both predictive accuracy and reliability. These results highlight the limitations of current temporal architectures for robust fare prediction in real-world noise conditions.~\autoref {fig:timesnet_denoised} \&~\autoref{fig:timesnet_noised} show the performance of Timesnet with noisy and denoised data.

\subsection{\textit{Model Comparison Summary}}

\textbf{1. Interpretation:}

\textit{XGBoost:} It demonstrates the highest resilience to noise injection due to its inherent tree-ensemble structure and built-in regularization capabilities. The model maintains stable performance across both clean and noisy datasets, showing robust calibration and generalization.

\textit{GAT:} It is susceptible to input noise, mainly due to the disruption of graph topologies and the absence of a denoising mechanism. This results in degraded calibration, elevated uncertainty, and reduced predictive stability under noisy conditions.

\textit{TimesNet:} It benefits moderately from denoising but remains less reliable than XGBoost in both calibration and OOD (out-of-distribution) robustness. Its performance suffers from architectural sensitivity to periodic sequences and higher memory consumption. 

~\autoref{tab:comp} presents an overall comparison between the models.

\vspace{2mm}
\begin{table}[H]
\centering
\resizebox{\columnwidth}{!}{%
\begin{tabular}{|l|c|c|c|c|c|c|}
\hline
\textbf{Model} & \textbf{MAE (Clean)} & \textbf{MSE (Clean)} & \textbf{R\textsuperscript{2} (Clean)} & \textbf{MAE (Noisy)} & \textbf{MSE (Noisy)} & \textbf{R\textsuperscript{2} (Noisy)} \\
\hline
\textbf{XGBoost} & 0.1040 & 0.0279 & 0.9910 & 0.8203 & 1.4407 & 0.7326 \\
\textbf{TimesNet} & 1.3549 & 3.3534 & -0.0779 & 1.7303 & 5.8835 & -0.0919 \\
\textbf{GAT} & 1.1057 & 2.4414 & 0.2153 & 2.5036 & 11.4375 & -1.1237 \\
\hline
\end{tabular}
}
\vspace{1mm}
\caption{Model Comparison}
\label{tab:comp}
\end{table}

\section{Conclusion}

This study presents a comprehensive empirical evaluation of three distinct modeling strategies—XGBoost, Graph Attention Networks (GAT), and TimesNet—applied to the task of taxi fare prediction using a large-scale, real-world dataset comprising over 55 million records. Both clean and synthetically noise-augmented variants of the dataset were utilized to benchmark each model's accuracy, robustness, and reliability.

The results indicate that XGBoost offers a strong balance between interpretability and performance. It achieves consistently low error rates across both clean and noisy inputs, exhibiting superior calibration and generalization. However, its dependence on manual feature engineering limits its scalability in dynamic or unstructured environments.

In contrast, GAT demonstrates the potential of relational learning by capturing spatial dependencies through graph structures. Although capable of modeling complex relationships, its performance is adversely impacted by noise, and it incurs high computational costs due to the requirements of attention-based processing and mini-batching.

TimesNet, a temporal deep learning model, effectively models global-local temporal patterns and demonstrates the highest tolerance to noise (MAE 1.35). However, it struggles with calibration and OOD generalization and is resource-intensive due to long sequence handling.

Through the integration of ensemble-based uncertainty estimation, calibration metrics, and bin-wise error analyses, this study highlights the significance of denoising mechanisms and robust architectural design for real-world deployment. Denoising autoencoders were found to enhance model stability under perturbations, mainly benefiting TimesNet.

In conclusion, the optimal model choice is contingent on deployment requirements:

\begin{itemize}
    \item For interpretable, low-latency predictions in structured tabular data environments, XGBoost is a suitable candidate.
    \item In spatially complex scenarios with relational dynamics, GAT remains a compelling direction, albeit with a need for improvements in robustness.
    \item For temporal forecasting in high-noise environments, TimesNet offers a practical solution, although its computational demands may limit its deployment.

\end{itemize}

\section{Future Work}
This study opens several avenues for future exploration. First, integrating external data sources, such as weather conditions, traffic congestion indices, and event calendars, may enhance model accuracy and context-awareness. Second, adversarial training and domain adaptation techniques can be employed to improve robustness to distributional shifts in real-world deployments. Third, deploying these models in real-time fare estimation engines and studying their latency, interpretability, and scalability trade-offs can offer valuable operational insights. Lastly, extending the analysis to multi-city or cross-regional taxi datasets may further validate the model's generalization and fairness across urban settings.


\end{document}